\documentclass[runningheads]{llncs}

\usepackage{hyperref}
\usepackage{xr-hyper}                  
\usepackage[T1]{fontenc}
\usepackage{graphicx}
\usepackage{color}

\usepackage{url}

\usepackage{graphicx}
\usepackage{algorithm}
\usepackage{algorithmic}
\usepackage{booktabs}
\usepackage[table]{xcolor}
\usepackage{colortbl}
\usepackage{xcolor}
\usepackage{pgf}
\usepackage{pgfmath}
\definecolor{MidnightBlue}{rgb}{0.1, 0.1, 0.44}

\usepackage{makecell} 
\usepackage{subcaption}
\usepackage{wrapfig}
\usepackage[authormarkup=none]{changes}
\definechangesauthor[name={Author}, color=black]{anon}
\usepackage{amsmath}
\usepackage{amssymb}

\usepackage{enumitem}
\setlist{nosep,leftmargin=*}

\begin{document}
\title{Distribution-Aware \texttt{Diffusion-LLM} for \\ Robust Ultra-Long-Term Time Series Forecasting}
\titlerunning{Diffusion-LLM for Ultra-Long-Term Forecasting}
%

\author{Falguni Ghosh\inst{1,3}\orcidID{0000-0002-2786-2044} \and
Vahid Hashemi\inst{3}\orcidID{0000-0002-9167-7417} \and
Bernhard Kainz\inst{1,2}\orcidID{0000-0002-7813-5023}}
\authorrunning{F. Ghosh et al.}
\institute{
Friedrich-Alexander-Universität Erlangen-Nürnberg, Erlangen, 91052, Germany\\
\email{\{falguni.ghosh\}@fau.de}
\and
Imperial College London, London SW7 2RH, United Kingdom
\and
AUDI AG, Auto-Union-Straße 1, 85057 Ingolstadt, Germany\\
}

\maketitle              
\begin{abstract}

Time series forecasting is a fundamental machine learning task. Recent work has explored Large Language Models (LLMs) for this purpose due to their strong generalization, pattern recognition, and zero-shot or few‑shot capabilities. Despite their suitability for long‑context learning, LLMs face challenges in multimodal settings: they lack calibrated probabilistic modeling for non‑text data and struggle to align heterogeneous representations. To address these issues, we propose a new framework \texttt{Diffusion‑LLM} that integrates a conditional diffusion model into an LLM-based forecasting pipeline. This joint design enables learning the conditional distribution of future data while improving semantic alignment in a shared latent space. We evaluate \texttt{Diffusion‑LLM} on six long-term forecasting benchmarks, including ETT, Weather, and ECL. Our method consistently outperforms existing LLM-based baseline, achieving notable gains in ultra‑long‑term and few‑shot forecasting and demonstrating the value of distribution-aware regularization for enhancing robustness and generalization in time series LLMs.

\end{abstract}
\section{Introduction}

Time series forecasting is essential in domains such as energy systems~\cite{9963650,CHOU2018709}, healthcare~\cite{10.1145/3531326}, climate science~\cite{KAREVAN20201}, and supply chain management~\cite{PACELLA2021604}. Many applications, including energy demand planning, climate modeling, and battery lifetime prediction~\cite{LI2024101891,WANG20231163} require ultra-long-term forecasts extending thousands or more steps ahead, often from limited historical data.

LLMs have recently emerged as promising forecasters due to their strong generalization, pattern recognition, and zero‑/few‑shot abilities~\cite{gruver_large_2023}. However, applying pretrained LLMs to time series remains challenging. Their representations are tuned for semantic structure in language, not temporal dynamics, making cross‑modal alignment difficult and leading to degraded performance and potential multimodal hallucinations~\cite{shukor2024implicit}. Moreover, MSE‑trained LLM forecasters tend to regress toward the mean and fail to capture the full distribution of possible futures, especially for irregular or noisy series~\cite{tang_time_2025}. As generation progresses, attention increasingly concentrates on recent predictions, reducing global context awareness and amplifying uncertainty underestimation~\cite{shi2023trustingevidencehallucinatecontextaware,10.1145/3703155}.

To address these issues, we incorporate a Denoising Diffusion Probabilistic Model (DDPM)~\cite{10.5555/3495724.3496298} into an LLM-based forecasting pipeline. Using the reprogramming strategy of TimeLLM~\cite{jin2024timellm}, both inputs and targets are embedded into a shared token space. The DDPM is jointly trained to estimate the conditional distribution of forecast embeddings given the lookback window, providing a distribution-aware signal that regularizes the LLM and strengthens multimodal alignment. This results in refinement of the shared embedding space, improved robustness, and long-horizon forecasting. Our key contributions are:

\begin{itemize}
\item We introduce DDPMs as implicit regularizers for multimodal LLMs, enabling joint alignment and distribution modeling in a unified embedding space.
\item We propose \texttt{Diffusion‑LLM}, a framework that models the distribution of reprogrammed time series patches to enhance temporal reasoning.
\item We show that our method significantly improves ultra‑long‑term and few‑shot forecasting performance across multiple benchmarks.
\end{itemize}

\section{Related Work}

\subsection{LLM in Time Series Forecasting:}

Recent research adapts LLMs to time series using several strategies. \textbf{Prompting-based methods} treat time series as raw text~\cite{xue2023promptcast,gruver_large_2023}, but lose temporal semantics due to modality mismatch. Quantization approaches discretize sequences via VQ-VAE or clustering~\cite{talukder2024totem,yu2023temporaldatameetsllm}, often requiring two‑stage training. \textbf{Vision‑as‑bridge methods} encode series as images interpreted by vision‑language models~\cite{fe3ce8147b534cbca5917d216231d2e0}, but rely on paired visual data and lack generality. \textbf{Tool‑augmented approaches} let LLMs generate code or API calls~\cite{qin2024toolllm} though they introduce complexity and are not end-to-end forecasters.

\textbf{Alignment-based approaches} instead learn time-series encodings compatible with LLM semantic spaces~\cite{zhang2024largelanguagemodelstime}. These methods fall into two categories:

\begin{itemize}
    \item \textbf{Contrastive alignment:} ETP~\cite{10446742}, TEST~\cite{sun2024test}, and TENT~\cite{zhou2023tentconnectlanguagemodels} use contrastive objectives to align sensor or physiological signals with text descriptions, effective when multimodal pairs exist.

    \item \textbf{LLM-backbone alignment:} GPT4TS~\cite{zhou2023one}, LLM4TS~\cite{10.1145/3719207}, and TimeLLM~\cite{jin2024timellm} feed reprogrammed time-series patches into frozen or partially frozen LLMs. GPT4TS freezes self‑attention layers to preserve pretrained knowledge; LLM4TS uses autoregressive alignment followed by parameter-efficient tuning; TimeLLM reprograms series into token sequences resembling natural language. Time‑VLM \cite{zhong2025timevlm} extends this paradigm to include image modalities.
\end{itemize}

Our work builds on this alignment-based direction but introduces diffusion-based regularization to strengthen distributional modeling, an aspect overlooked in classical models~\cite{benidis2022deep,yang2025miel,3b1355aedd1041f1853e609a410576f3,wen2018multihorizonquantilerecurrentforecaster} which focus on deterministic or multiscale decomposition rather than probabilistic uncertainty or multimodal alignment.

\begin{figure*}[bt]
\centering
\includegraphics[width=0.75\textwidth]{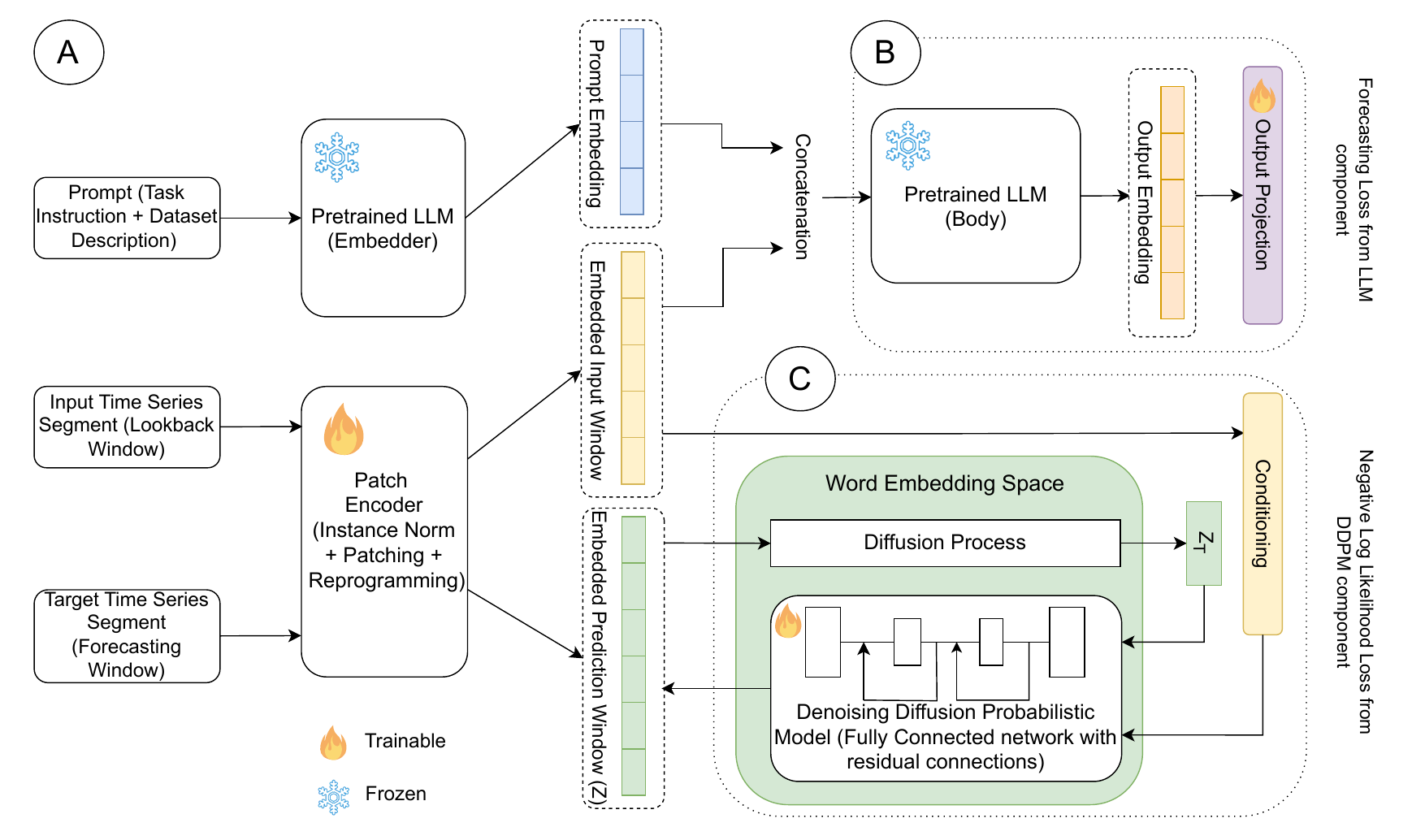} 
\caption{Training architecture of \texttt{Diffusion-LLM}. (A) The prompt, input, and target time series are reprogrammed into a shared token embedding space using a frozen LLM encoder and a trainable patch encoder. (B) The encoded input is used for direct forecasting via a frozen LLM body + trainable output projection module. (C) A conditional DDPM is trained to model the distribution of the encoded target, conditioned on the input, by predicting the added noise. The final loss combines forecasting and diffusion-based regularization.}
\label{fig_train}
\end{figure*}

\subsection{DDPM in Time Series Forecasting:}

DDPM-based forecasters generally pair diffusion models with autoregressive backbones. TimeGrad~\cite{pmlr-v139-rasul21a} corrupts future values with noise and denoises them conditioned on RNN‑encoded lookback windows. ScoreGrad~\cite{yan2021scoregradmultivariateprobabilistictime} follows a similar feature extraction pipeline but employs conditional SDE-based score matching. Unlike these models, we do not use DDPMs as stand‑alone generative forecasters. Instead, we leverage DDPMs as auxiliary learners that regularize LLM-based predictors, improving robustness and uncertainty modeling without replacing the LLM’s forecasting role.

\section{Methodology}

Our proposed framework, \texttt{Diffusion-LLM}, enhances LLM-based time series forecasting by integrating a conditional DDPM as a regularizer. The model estimates the conditional distribution of the forecast window given the lookback window within a shared embedding space of text prototypes produced through time‑series reprogramming, improving both probabilistic modeling and multimodal alignment. An overview of the training architecture is shown in Figure~\ref{fig_train}. The framework consists of three main components:

\vspace{0.25em}

\noindent\textbf{\textit{A. Time Series Encoder (Reprogramming and Embedding):}}

\begin{figure}[tb]
\centering
\begin{subfigure}[t]{0.25\textwidth}
    \centering
    \includegraphics[width=\linewidth]{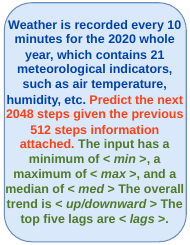}
    \caption{Prompt.}
    \label{fig_prompt}
\end{subfigure}
\hfill
\begin{subfigure}[t]{0.33\textwidth}
    \centering
    \includegraphics[width=\linewidth]{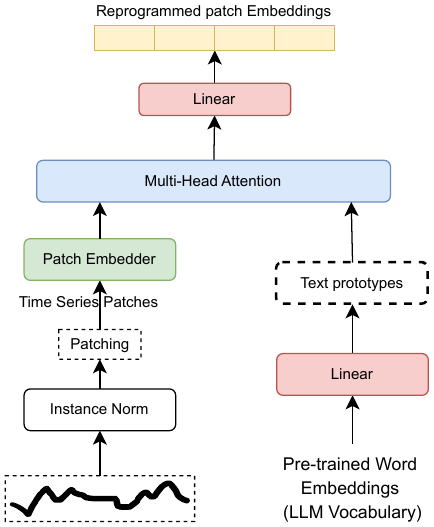}
    \caption{Patch Encoding.}
    \label{fig_reprogram}
\end{subfigure}
\hfill
\begin{subfigure}[t]{0.34\textwidth}
    \centering
    \includegraphics[width=\linewidth]{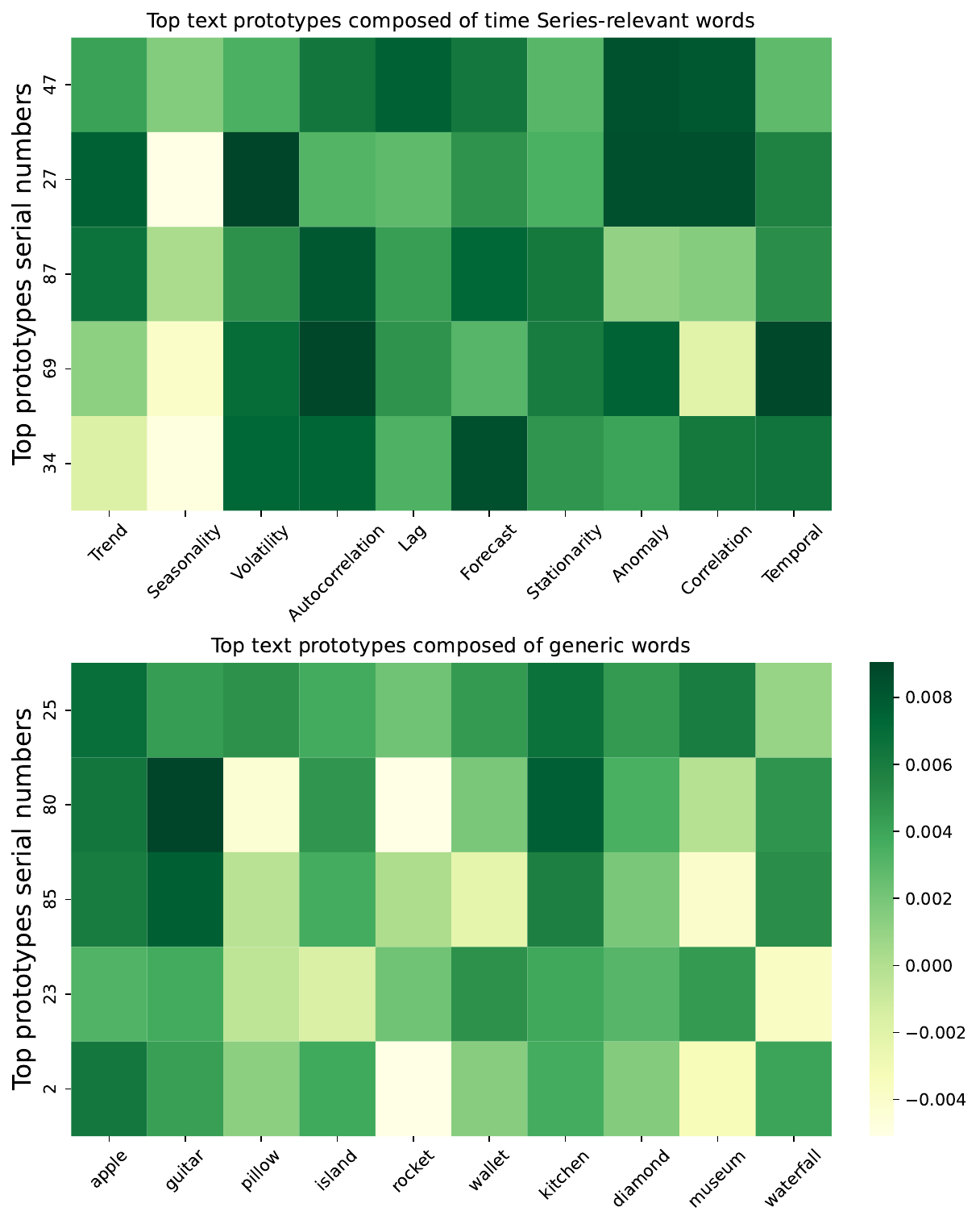}
    \caption{Prototype composition.}
    \label{fig_proto_compose}
\end{subfigure}
\caption{(a) Sample prompt used for the Weather dataset. 
(b) Patch encoding mechanism, where a time series window is normalized, patched, and matched to text prototypes learned and derived from pretrained LLM vocabulary. Attention between patches and prototypes enables the model to associate temporal patterns with language-like cues (`steady up', `periodic pattern' etc.). 
(c) Prototype composition visualization, showing the top-matched prototypes for a set of time-series-related and generic words after training, illustrating how the learned prototype space can form prototypes with slightly higher affinity for time series-related vocabulary over generic words in initial linear layer already. This distinction becomes more pronounced after the attention layers (Figure~\ref{fig:proto_attention}).}
\label{fig:combined_patch_reprog}
\end{figure}

Following the TimeLLM reprogramming strategy~\cite{jin2024timellm}, raw time series is tokenized into patches and mapped to text-like prototypes via an attention-based mechanism (Figure~\ref{fig_reprogram}). During training, both the lookback window $x$ and forecast window $y$ are encoded using the shared encoder $\phi_{\text{llmenc}}$, producing latent representations $z_x$ and $z_y$. We maintain the three-part prompt design from the baseline - dataset details, task instruction, and statistical information (Figure~\ref{fig_prompt}). For simplicity, we omit the frozen prompt embedder in notation and denote the patch encoder directly as $\phi_{\text{llmenc}}$:
\begin{equation}
z_x = \phi_{\text{llmenc}}(x), \quad z_y = \phi_{\text{llmenc}}(y).
\end{equation}

The encoded time series becomes a semantic representation (prototype) describing its patterns and statistics in the LLM’s embedding space (Figure~\ref{fig_proto_compose}), enabling the pretrained model to process it using its native architecture.

\vspace{0.25em}

\noindent\textbf{\textit{B. Forecasting via LLM:}}

The encoded input $z_x$ is passed to an output module $\phi_{\text{llmout}}$, consisting of the pretrained frozen LLM body and a small trainable output projection layer, to produce the forecast $\hat{y}$:
\begin{equation}
\hat{y} = \phi_{\text{llmout}}(z_x), \quad 
\mathcal{L}_{\text{forecast}} = \| y - \hat{y} \|^2.
\end{equation}
This component exploits the pretrained reasoning and pattern-recognition capabilities of LLMs without full finetuning, retaining efficiency and generalization. As shown in~\cite{jin2024timellm,dombrowski2024trade}, reprogramming frozen LLMs can be more efficient than parameter‑efficient finetuning approaches such as QLoRA~\cite{dettmers2023qlora}.


\begin{algorithm}[htbp]
\caption{\textbf{Diffusion-LLM Training}}
\label{alg:training}
\begin{algorithmic}
\REQUIRE Time series dataset $\mathcal{D} = \{(x, y)\}$, LLM encoder module $\phi_{\text{llmenc}}$, LLM output module $\phi_{\text{llmout}}$, DDPM model $\theta_{\text{ddpm}}$, regularization weight $\lambda$.
Let $\alpha_t$ denote the noise schedule coefficients at time t and $\bar{\alpha}_t = \prod_{s=1}^{t} \alpha_s$.
\STATE Initialize parameters of $\phi_{\text{llmenc}}, \phi_{\text{llmout}}, \theta_{\text{ddpm}}$.
\FOR{each training iteration}
    \STATE Sample a batch $\mathcal{B} = \{(x_i, y_i)\}$ from $\mathcal{D}$
    \FOR{each $(x, y)$ in $\mathcal{B}$}
        \STATE \textbf{1. Encode input and target windows}
        \STATE \hspace{1em} (a) $z_x \leftarrow \phi_{\text{llmenc}}(x)$, (b) $z_y \leftarrow \phi_{\text{llmenc}}(y)$

        \STATE \textbf{2. Forecasting prediction and loss}
        \STATE \hspace{1em} (a) $\hat{y} \leftarrow \phi_{\text{llmout}}(z_x)$, (b) $\mathcal{L}_{\text{forecast}} \leftarrow \| y - \hat{y} \|^2$

        \STATE \textbf{3. DDPM loss}
        \STATE \hspace{1em} (a) Sample noise $\epsilon \sim \mathcal{N}(0, I)$ and timestep $t \sim \text{Uniform}(1, T)$
        \STATE \hspace{1em} (b) Noised sample: $\tilde{z}_y = \sqrt{\bar{\alpha}_t} z_y + \sqrt{1 - \bar{\alpha}_t} \cdot \epsilon$
        \STATE \hspace{1em} (c) Predict noise: $\hat{\epsilon} \leftarrow \theta_{\text{ddpm}}(\tilde{z}_y, t, z_x)$
        \STATE \hspace{1em} (d) $\mathcal{L}_{\text{ddpm}} \leftarrow \| \epsilon - \hat{\epsilon} \|^2$

        \STATE \textbf{4. Combine losses}
        \STATE \hspace{1em} (a) $\mathcal{L}_{\text{joint}} \leftarrow \mathcal{L}_{\text{forecast}} + \lambda \cdot \mathcal{L}_{\text{ddpm}}$
    \ENDFOR
    \STATE Update $\phi_{\text{llmenc}}$, trainable projection part of $\phi_{\text{llmout}}$, $\theta_{\text{ddpm}}$ using gradients of $\mathcal{L}_{\text{joint}}$
\ENDFOR
\end{algorithmic}
\end{algorithm}

\vspace{0.25em}

\noindent\textbf{\textit{C. Distribution Regularization via DDPM:}}

To strengthen the model's ability to capture the token distribution of time series representations, we incorporate a conditional DDPM that learns the conditional distribution $p(z_y \mid z_x)$ through a denoising process. During training, noise is added to $z_y$ to obtain $\tilde{z}_y$, and the DDPM predicts the noise $\epsilon$:
\begin{equation}
\tilde{z}_y \sim q(\tilde{z}_y \mid z_y, t), \quad 
\hat{\epsilon} = \epsilon_\theta(\tilde{z}_y, t, z_x), \quad
\mathcal{L}_{\text{ddpm}} = \| \epsilon - \hat{\epsilon} \|^2.
\end{equation}
Here, $\epsilon_\theta$ denotes the DDPM denoising network parameterized by $\theta_{\text{ddpm}}$, $t$ is the diffusion timestep, and $\epsilon \sim \mathcal{N}(0,I)$ is the sampled noise in the forward noising process $q(\tilde{z}_y \mid z_y, t)$. As shown in~\cite{10.5555/3495724.3496298}, this objective is equivalent to maximizing the conditional likelihood,
\begin{equation}
\mathcal{L}_{\text{ddpm}} = -\log p_\theta(z_y \mid z_x).
\end{equation}
The DDPM thus serves as both a probabilistic constraint and an auxiliary learner that enriches the shared embedding space through semantic alignment. The overall model is jointly optimized for forecasting and distribution estimation via
\begin{equation}
\mathcal{L}_{\text{joint}} 
= \mathcal{L}_{\text{forecast}} 
+ \lambda \cdot \mathcal{L}_{\text{ddpm}}.
\end{equation}
This regularization is alignment-agnostic and can be integrated into existing LLM-based forecasting methods with minimal modification, as demonstrated in our enhancement of TimeLLM~\cite{jin2024timellm}. We jointly optimize all the learnable parameters. The full training procedure is provided in Algorithm~\ref{alg:training}.

\begin{wrapfigure}{r}{0.6\textwidth}  
    \centering
    \vspace{10pt} 
    \includegraphics[width=0.5\textwidth]{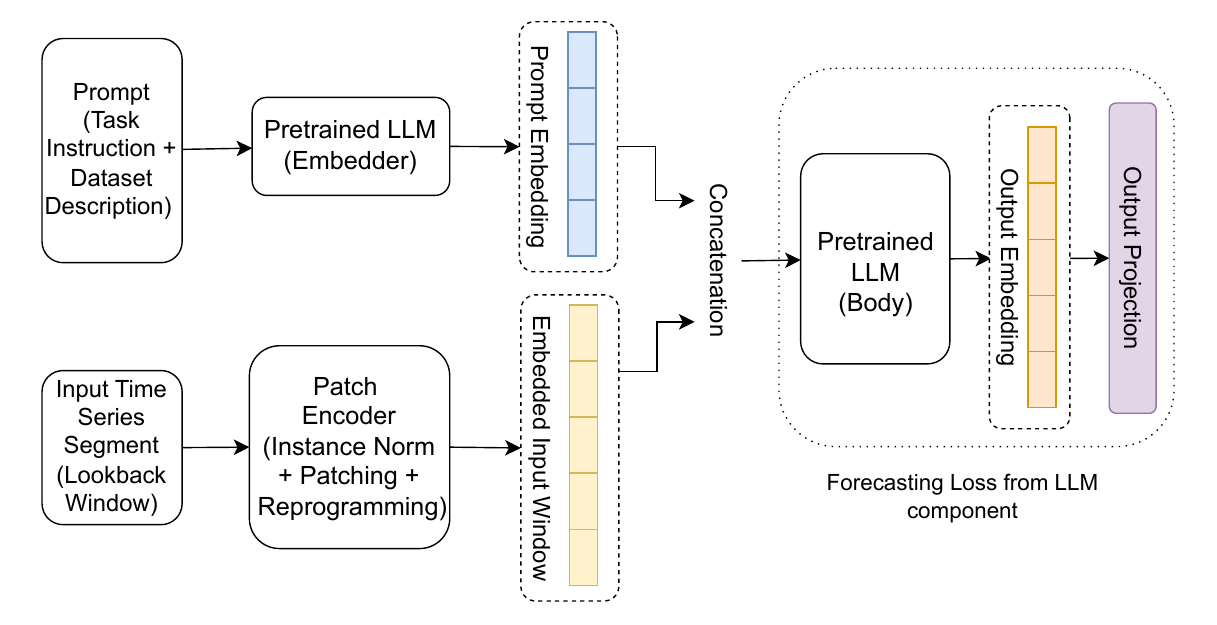}
    \caption{Inference pipeline of \texttt{Diffusion-LLM}. Only the LLM modules are used to generate forecasts for new inputs.}
    \label{fig_infer}
    \vspace{10pt} 
\end{wrapfigure}

During inference, only the LLM modules are used to generated forecasts (Figure~\ref{fig_infer}). The inference steps are formally defined in  Algorithm~\ref{alg:inference}.

\begin{algorithm}[htbp]
\caption{\textbf{Diffusion-LLM Inference}}
\label{alg:inference}
\begin{algorithmic}
\REQUIRE Input time series $x$, trained encoder $\phi_{\text{llmenc}}$, trained output module $\phi_{\text{llmout}}$
\STATE \textbf{1. Encode the input window}
\STATE \hspace{1em} $z_x \leftarrow \phi_{\text{llmenc}}(x)$
\STATE \textbf{2. Generate forecast}
\STATE \hspace{1em} $\hat{y} \leftarrow \phi_{\text{llmout}}(z_x)$
\RETURN $\hat{y}$ as the predicted forecast window
\end{algorithmic}
\end{algorithm}

\section{Experiments and Results}

\noindent\textbf{\textit{A. Model Architecture:}}
We use the 7B variant of LLaMA~\cite{touvron2023llamaopenefficientfoundation} as the backbone LLM. The diffusion module is a lightweight DDPM implemented as a stack of fully connected layers with skip connections (details in~\ref{subsec:expt_details}). All experiments are run on NVIDIA A100 and H100 GPUs.

\noindent\textbf{\textit{B. Long-Term Forecasting:}}
We evaluate \texttt{Diffusion-LLM} on six standard long-term forecasting benchmarks: ETTh1, ETTh2, ETTm1, ETTm2~\cite{haoyietal-informer-2021}, Weather, and ECL~\cite{wu2023timesnet} (details in~\ref{subsec:dataset_details}). As our method serves as an add-on to existing LLM-based approaches, we report competitive results with established benchmarks (Table~\ref{tab:forecasting_results_reordered}) and provide a comprehensive comparison against the TimeLLM baseline, including mean and standard deviation for MSE and MAE metrics (Table~\ref{tab:full_ltf_comparison}, Table~\ref{tab:full_fewshot_comparison}). For long-term forecasting, \texttt{Diffusion-LLM} achieves performance comparable to TimeLLM.

\newcommand{\spc}{0.01cm}

\begin{table*}[t]
\centering
\footnotesize
\renewcommand{\arraystretch}{1.3}
\setlength{\tabcolsep}{3pt}
\caption{Comparison of TimeLLM~\cite{jin2024timellm} and \texttt{Diffusion-LLM} on long-term and ultra-long-term forecasting across standard benchmarks. \textbf{Long-term} results are averaged over horizons $H \in \{96, 192, 336, 720\}$ with a 512-step input; \textbf{ultra-long-term} averages over $H \in \{1024, 2048\}$. Each cell reports MSE, MAE, and standard deviations across runs (lower is better; best in bold). \texttt{Diffusion-LLM} outperforms TimeLLM on 4/6 datasets in the ultra-long-term setting, with substantial gains on more challenging smaller datasets such as ETTh1 and ETTh2.}
\resizebox{\textwidth}{!}{%
\begin{tabular}{l p{\spc} cc p{\spc} cc p{\spc} cc p{\spc} cc}
\toprule
\textbf{Dataset} 
&& \multicolumn{5}{c}{\textbf{Long-term}} 
&& \multicolumn{5}{c}{\textbf{Ultra-long-term}} \\
 && \multicolumn{2}{c}{\textbf{TimeLLM}} && \multicolumn{2}{c}{\textbf{Diffusion-LLM (Ours)}} 
 && \multicolumn{2}{c}{\textbf{TimeLLM}} && \multicolumn{2}{c}{\textbf{Diffusion-LLM (Ours)}} \\
 && MSE & MAE && MSE & MAE && MSE & MAE && MSE & MAE \\
\midrule
ETTh1 && $0.449_{\pm 0.025}$ & $0.457_{\pm 0.015}$ && $\textbf{0.427}_{\pm 0.004}$ & $\textbf{0.446}_{\pm 0.010}$ && $0.758_{\pm 0.018}$ & $0.600_{\pm 0.011}$ && $\textbf{0.612}_{\pm 0.011}$ & $\textbf{0.558}_{\pm 0.004}$ \\
ETTh2 && $\textbf{0.373}_{\pm 0.009}$ & $\textbf{0.409}_{\pm 0.006}$ &&  $0.387_{\pm 0.003}$ & $0.425_{\pm 0.002}$ &&  $0.589_{\pm 0.013}$ & $0.543_{\pm 0.007}$ &&  $\textbf{0.522}_{\pm 0.009}$ & $\textbf{0.512}_{\pm 0.004}$ \\
ETTm1 && $0.381_{\pm 0.008}$ & $0.406_{\pm 0.006}$ &&  $\textbf{0.376}_{\pm 0.004}$ & $\textbf{0.399}_{\pm 0.002}$ &&  $0.484_{\pm 0.009}$ & $0.472_{\pm 0.012}$ &&  $\textbf{0.465}_{\pm 0.001}$ & $\textbf{0.452}_{\pm 0.001}$ \\
ETTm2 && $\textbf{0.271}_{\pm 0.003}$ & $\textbf{0.330}_{\pm 0.003}$ && $0.334_{\pm 0.003}$ & $0.369_{\pm 0.001}$ && $\textbf{0.410}_{\pm 0.020}$ & $\textbf{0.425}_{\pm 0.014}$ && $0.422_{\pm 0.008}$ & $0.436_{\pm 0.004}$ \\
Weather && $\textbf{0.259}_{\pm 0.019}$ & $\textbf{0.288}_{\pm 0.017}$ && $0.304_{\pm 0.001}$ & $0.329_{\pm 0.001}$ && $0.424_{\pm 0.008}$ & $0.401_{\pm 0.004}$ && $\textbf{0.407}_{\pm 0.001}$ & $\textbf{0.394}_{\pm 0.001}$ \\
ECL && $\textbf{0.171}_{\pm 0.002}$ & $\textbf{0.277}_{\pm 0.003}$ && $0.200_{\pm 0.004}$ & $0.303_{\pm 0.002}$ && $\textbf{0.272}_{\pm 0.001}$ & $\textbf{0.356}_{\pm 0.000}$ && $0.297_{\pm 0.005}$ & $0.376_{\pm 0.004}$ \\
\bottomrule
\end{tabular}
}
\label{tab:full_ltf_comparison}
\end{table*}

\begin{table*}[tb]
\centering
\footnotesize
\renewcommand{\arraystretch}{1.3}
\setlength{\tabcolsep}{3pt}
\caption{Comparison of TimeLLM~\cite{jin2024timellm} and \texttt{Diffusion-LLM} in few-shot long and ultra-long-term forecasting across standard benchmarks. \textbf{Few-shot (10\% or 5\%)} indicate training with only 10\% or 5\% of the available data to assess generalization under scarcity. Other experimental details follow the protocol of Table~\ref{tab:full_ltf_comparison}. ‘–’ denotes insufficient data for a meaningful training set. \texttt{Diffusion-LLM} consistently outperforms TimeLLM in few-shot ultra-long-term settings.}
\resizebox{\textwidth}{!}{%
\begin{tabular}{l p{\spc} cc p{\spc} cc p{\spc} cc p{\spc} cc p{\spc} cc p{\spc} cc p{\spc} cc p{\spc}cc}
\toprule
\textbf{Dataset} 
&& \multicolumn{5}{c}{\textbf{Few-shot (10\%) long-term}} 
&& \multicolumn{5}{c}{\textbf{Few-shot (10\%) ultra-long-term}} 
&& \multicolumn{5}{c}{\textbf{Few-shot (5\%) long-term}} 
&& \multicolumn{5}{c}{\textbf{Few-shot (5\%) ultra-long-term}} \\
 && \multicolumn{2}{c}{\textbf{TimeLLM}} && \multicolumn{2}{c}{\textbf{Diffusion-LLM (Ours)}} 
 && \multicolumn{2}{c}{\textbf{TimeLLM}} && \multicolumn{2}{c}{\textbf{Diffusion-LLM (Ours)}} 
 && \multicolumn{2}{c}{\textbf{TimeLLM}} && \multicolumn{2}{c}{\textbf{Diffusion-LLM (Ours)}} 
 && \multicolumn{2}{c}{\textbf{TimeLLM}} && \multicolumn{2}{c}{\textbf{Diffusion-LLM (Ours)}} \\
 && MSE & MAE && MSE & MAE && MSE & MAE && MSE & MAE && MSE & MAE && MSE & MAE && MSE & MAE && MSE & MAE \\
\midrule
ETTh1 && $0.834_{\pm 0.073}$ & $0.614_{\pm 0.022}$ && $\textbf{0.662}_{\pm 0.004}$ & $\textbf{0.564}_{\pm 0.001}$ && - & - && - & - && $0.988_{\pm 0.066}$ & $0.662_{\pm 0.021}$ && $\textbf{0.728}_{\pm 0.029}$ & $\textbf{0.582}_{\pm 0.013}$ && - & - && - & - \\
ETTh2 && $0.422_{\pm 0.009}$ & $0.443_{\pm 0.005}$ && $\textbf{0.398}_{\pm 0.003}$ & $\textbf{0.432}_{\pm 0.002}$ && - & - && - & - && $0.415_{\pm 0.014}$ & $0.435_{\pm 0.008}$ && $\textbf{0.392}_{\pm 0.003}$ & $\textbf{0.428}_{\pm 0.003}$ && - & - && - & - \\
ETTm1 && $0.504_{\pm 0.001}$ & $\textbf{0.462}_{\pm 0.003}$ && $\textbf{0.502}_{\pm 0.029}$ & $0.464_{\pm 0.014}$ && $1.056_{\pm 0.101}$ & $0.691_{\pm 0.036}$ && $\textbf{0.660}_{\pm 0.062}$ & $\textbf{0.550}_{\pm 0.026}$ && $0.600_{\pm 0.011}$ & $0.515_{\pm 0.006}$ && $\textbf{0.528}_{\pm 0.014}$ & $\textbf{0.480}_{\pm 0.005}$ && $0.924_{\pm 0.032}$ & $0.666_{\pm 0.011}$ && $\textbf{0.628}_{\pm 0.003}$ & $\textbf{0.536}_{\pm 0.001}$ \\
ETTm2 && $\textbf{0.327}_{\pm 0.017}$ & $\textbf{0.361}_{\pm 0.008}$ && $0.336_{\pm 0.003}$ & $0.370_{\pm 0.003}$ && $0.582_{\pm 0.022}$ & $0.506_{\pm 0.003}$ && $\textbf{0.442}_{\pm 0.000}$ & $\textbf{0.447}_{\pm 0.000}$ && $\textbf{0.330}_{\pm 0.005}$ & $\textbf{0.367}_{\pm 0.003}$ && $0.346_{\pm 0.001}$ & $0.381_{\pm 0.003}$ && $0.522_{\pm 0.018}$ & $0.480_{\pm 0.003}$ && $\textbf{0.450}_{\pm 0.004}$ & $\textbf{0.444}_{\pm 0.006}$ \\
Weather && $\textbf{0.256}_{\pm 0.000}$ & $\textbf{0.291}_{\pm 0.002}$ && $0.319_{\pm 0.008}$ & $0.340_{\pm 0.004}$ && $0.480_{\pm 0.005}$ & $0.430_{\pm 0.002}$ && $\textbf{0.428}_{\pm 0.003}$ & $\textbf{0.406}_{\pm 0.000}$ && $\textbf{0.304}_{\pm 0.006}$ & $\textbf{0.326}_{\pm 0.003}$ && $0.329_{\pm 0.005}$ & $0.347_{\pm 0.003}$ && $0.477_{\pm 0.007}$ & $0.434_{\pm 0.004}$ && $\textbf{0.424}_{\pm 0.006}$ & $\textbf{0.406}_{\pm 0.004}$ \\
ECL && $\textbf{0.190}_{\pm 0.000}$ & $\textbf{0.288}_{\pm 0.001}$ && $0.197_{\pm 0.000}$ & $0.294_{\pm 0.000}$ && $0.292_{\pm 0.000}$ & $0.367_{\pm 0.003}$ && $\textbf{0.281}_{\pm 0.001}$ & $\textbf{0.358}_{\pm 0.004}$ && $\textbf{0.192}_{\pm 0.000}$ & $\textbf{0.289}_{\pm 0.001}$ && $0.201_{\pm 0.003}$ & $0.298_{\pm 0.000}$ && - & - && - & - \\
\bottomrule
\end{tabular}
} 
\label{tab:full_fewshot_comparison}
\end{table*}

\begin{figure}[tb]
\centering
\resizebox{0.9\linewidth}{!}{%
\begin{minipage}[t]{0.31\textwidth}
    \vspace{0pt} 
    \centering
    \begin{subfigure}[t]{\textwidth}
        \centering
        \includegraphics[width=\linewidth]{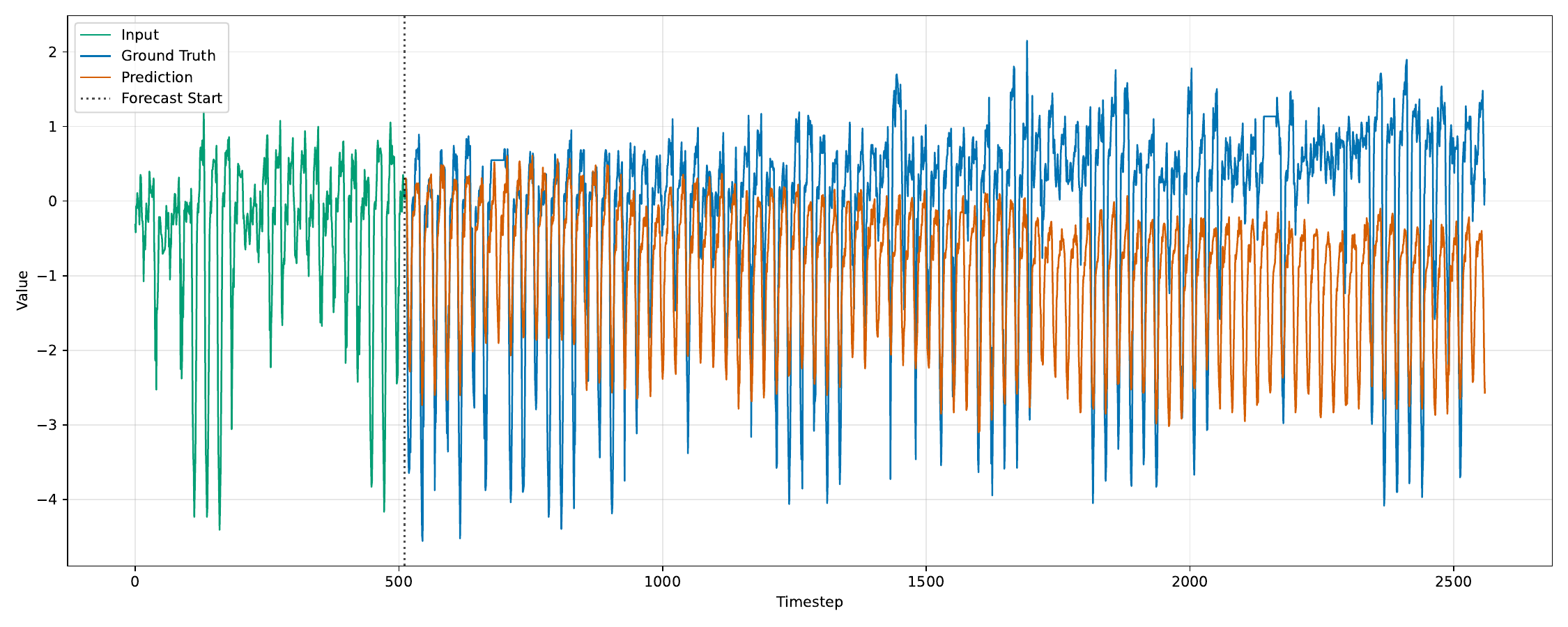}
        \caption*{TimeLLM}
        \label{fig:timeLLM}
    \end{subfigure}
    \vspace{0.5em}
    \begin{subfigure}[t]{\textwidth}
        \centering
        \includegraphics[width=\linewidth]{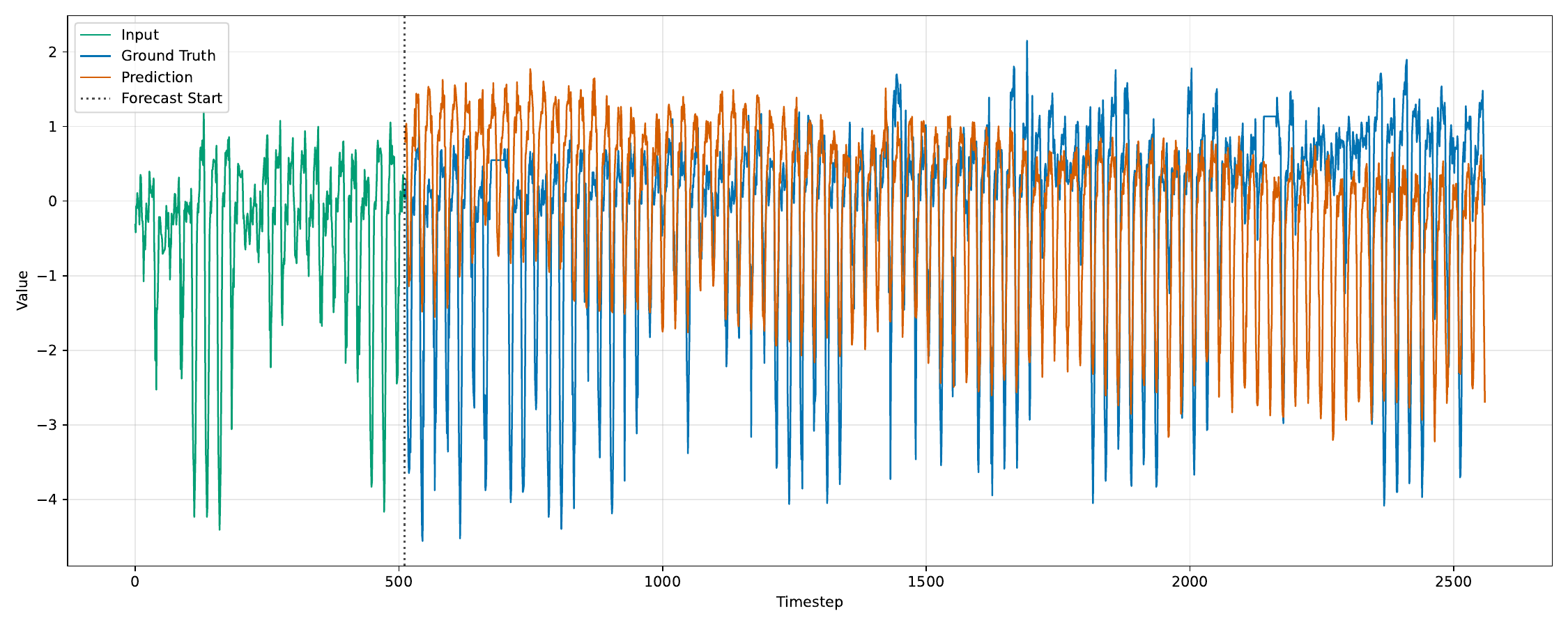}
        \caption*{Diffusion-LLM}
        \label{fig:diffusionLLM}
    \end{subfigure}
    \caption*{\textbf{(a)} Forecasting instance}
\end{minipage}
\hfill
\begin{minipage}[t]{0.54\textwidth}
    \vspace{0pt} 
    \centering
    \includegraphics[width=\linewidth]{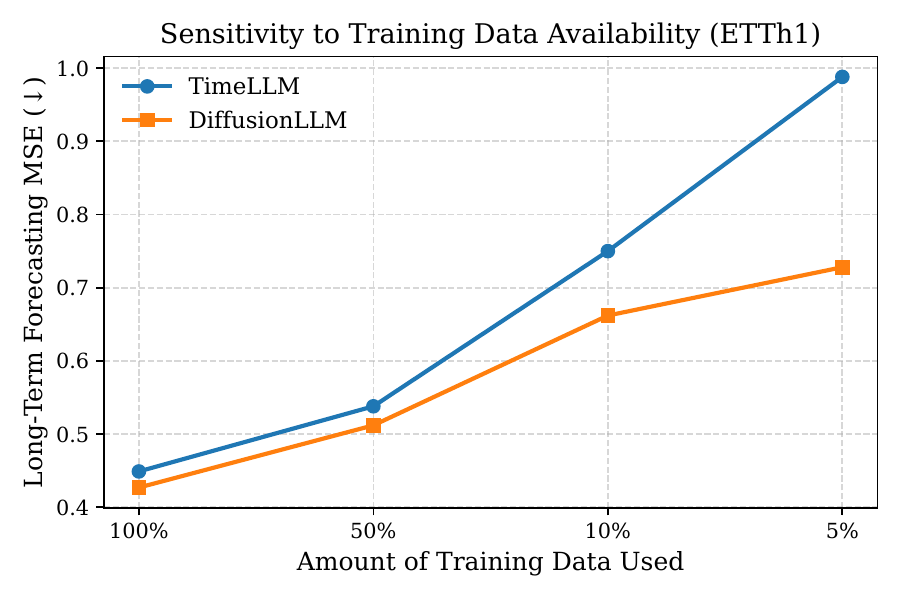}
    \caption*{\textbf{(b)} Performance under data-scarcity}
    \label{fig_fewshot_compare}
\end{minipage}
}
\caption{(a) Ultra-long-term forecasting on an ETTh1 sample (512 lookback, 2048 forecast). TimeLLM deviates faster than \texttt{Diffusion-LLM} in the later regions. (b) Long-term forecasting comparison under data-scarcity, showing slower degradation and greater robustness of \texttt{Diffusion-LLM}. Protocol follows Table~\ref{tab:full_ltf_comparison}.}
\label{fig:side_by_side_combined}
\end{figure}

\noindent\textbf{\textit{C. Ultra-Long-Term Forecasting:}}
Ultra-long-term forecasting remains highly challenging due to increased uncertainty and weaker dependence on recent history. We evaluate $\{1024, 2048\}$ forecast lengths on the same datasets, and our \texttt{Diffusion-LLM} consistently outperforms TimeLLM in this regime (Table~\ref{tab:full_ltf_comparison}). Improvements are especially pronounced on smaller datasets such as ETTh1 and ETTh2, with MSE reductions of 19.26\% and 11.38\%, respectively, highlighting the benefit of modeling the full conditional distribution of the target window.

\noindent\textbf{\textit{D. Few-Shot Forecasting (10\% and 5\%):}}
To assess few-shot generalization, we train both models using only 10\% of the data. \texttt{Diffusion-LLM} consistently surpasses TimeLLM across all long- and ultra-long‑term settings (Table~\ref{tab:full_fewshot_comparison}); on ETTh1, it achieves a 20.62\% improvement even for long-term horizons. In the more extreme 5\% setting, the advantage becomes even clearer, with \texttt{Diffusion-LLM} improving ETTh1 performance by 25.79\%. These results show that diffusion-based regularization substantially enhances generalization in low-data regimes without modifying the frozen LLM backbone.

Our results indicate that while multimodal LLMs can capture short-term structure, very long horizons with higher uncertainty require explicit distribution modeling. Incorporating the DDPM helps the encoder learn richer representations in the shared embedding space. As diffusion models estimate full probability distributions, the regularization they provide is most beneficial under high uncertainty; this can introduce a mild trade-off, slightly reducing point accuracy for short horizons. Empirically (Tables~\ref{tab:full_ltf_comparison}, \ref{tab:full_fewshot_comparison}), the gains are largest on challenging smaller datasets (e.g., ETTh1, ETTm1). Even for other datasets, benefits emerge as forecast horizons lengthen or training data becomes scarce. Figure~\ref{fig:side_by_side_combined} further shows that under severe data scarcity, \texttt{Diffusion-LLM} degrades substantially less than TimeLLM, underscoring its robustness in high-uncertainty settings.


\definecolor{lightblue1}{rgb}{0.95,0.98,1}
\definecolor{lightblue2}{rgb}{0.2,0.4,0.8}


\newcommand{\minvalETThOneMSE}{0.405}
\newcommand{\maxvalETThOneMSE}{1.040}
\newcommand{\minvalETThOneMAE}{0.420}
\newcommand{\maxvalETThOneMAE}{0.805}

\newcommand{\minvalETThTwoMSE}{0.330}
\newcommand{\maxvalETThTwoMSE}{6.736}
\newcommand{\minvalETThTwoMAE}{0.379}
\newcommand{\maxvalETThTwoMAE}{2.191}

\newcommand{\minvalETTmOneMSE}{0.347}
\newcommand{\maxvalETTmOneMSE}{0.961}
\newcommand{\minvalETTmOneMAE}{0.377}
\newcommand{\maxvalETTmOneMAE}{0.734}

\newcommand{\minvalETTmTwoMSE}{0.248}
\newcommand{\maxvalETTmTwoMSE}{1.479}
\newcommand{\minvalETTmTwoMAE}{0.311}
\newcommand{\maxvalETTmTwoMAE}{0.915}

\newcommand{\minvalWeatherMSE}{0.224}
\newcommand{\maxvalWeatherMSE}{0.803}
\newcommand{\minvalWeatherMAE}{0.263}
\newcommand{\maxvalWeatherMAE}{0.656}

\newcommand{\minvalECLMSE}{0.161}
\newcommand{\maxvalECLMSE}{0.338}
\newcommand{\minvalECLMAE}{0.252}
\newcommand{\maxvalECLMAE}{0.422}

\newcommand{\minvalILIMSE}{1.443}
\newcommand{\maxvalILIMSE}{7.382}
\newcommand{\minvalILIMAE}{0.797}
\newcommand{\maxvalILIMAE}{2.003}


\newcommand{\heatcellETThOneMSE}[1]{%
  \pgfmathtruncatemacro{\shade}{round(100*(#1 - \minvalETThOneMSE)/max(\maxvalETThOneMSE - \minvalETThOneMSE, 0.0001))}%
  \xdef\colorstring{lightblue1!\shade!lightblue2}%
  \cellcolor{\colorstring}{#1}%
}

\newcommand{\heatcellETThOneMAE}[1]{%
  \pgfmathtruncatemacro{\shade}{round(100*(#1 - \minvalETThOneMAE)/max(\maxvalETThOneMAE - \minvalETThOneMAE, 0.0001))}%
  \xdef\colorstring{lightblue1!\shade!lightblue2}%
  \cellcolor{\colorstring}{#1}%
}

\newcommand{\heatcellETThTwoMSE}[1]{%
  \pgfmathtruncatemacro{\shade}{round(100*(#1 - \minvalETThTwoMSE)/max(\maxvalETThTwoMSE - \minvalETThTwoMSE, 0.0001))}%
  \xdef\colorstring{lightblue1!\shade!lightblue2}%
  \cellcolor{\colorstring}{#1}%
}

\newcommand{\heatcellETThTwoMAE}[1]{%
  \pgfmathtruncatemacro{\shade}{round(100*(#1 - \minvalETThTwoMAE)/max(\maxvalETThTwoMAE - \minvalETThTwoMAE, 0.0001))}%
  \xdef\colorstring{lightblue1!\shade!lightblue2}%
  \cellcolor{\colorstring}{#1}%
}

\newcommand{\heatcellETTmOneMSE}[1]{%
  \pgfmathtruncatemacro{\shade}{round(100*(#1 - \minvalETTmOneMSE)/max(\maxvalETTmOneMSE - \minvalETTmOneMSE, 0.0001))}%
  \xdef\colorstring{lightblue1!\shade!lightblue2}%
  \cellcolor{\colorstring}{#1}%
}
\newcommand{\heatcellETTmOneMAE}[1]{%
  \pgfmathtruncatemacro{\shade}{round(100*(#1 - \minvalETTmOneMAE)/max(\maxvalETTmOneMAE - \minvalETTmOneMAE, 0.0001))}%
  \xdef\colorstring{lightblue1!\shade!lightblue2}%
  \cellcolor{\colorstring}{#1}%
}

\newcommand{\heatcellETTmTwoMSE}[1]{%
  \pgfmathtruncatemacro{\shade}{round(100*(#1 - \minvalETTmTwoMSE)/max(\maxvalETTmTwoMSE - \minvalETTmTwoMSE, 0.0001))}%
  \xdef\colorstring{lightblue1!\shade!lightblue2}%
  \cellcolor{\colorstring}{#1}%
}
\newcommand{\heatcellETTmTwoMAE}[1]{%
  \pgfmathtruncatemacro{\shade}{round(100*(#1 - \minvalETTmTwoMAE)/max(\maxvalETTmTwoMAE - \minvalETTmTwoMAE, 0.0001))}%
  \xdef\colorstring{lightblue1!\shade!lightblue2}%
  \cellcolor{\colorstring}{#1}%
}

\newcommand{\heatcellWeatherMSE}[1]{%
  \pgfmathtruncatemacro{\shade}{round(100*(#1 - \minvalWeatherMSE)/max(\maxvalWeatherMSE - \minvalWeatherMSE, 0.0001))}%
  \xdef\colorstring{lightblue1!\shade!lightblue2}%
  \cellcolor{\colorstring}{#1}%
}
\newcommand{\heatcellWeatherMAE}[1]{%
  \pgfmathtruncatemacro{\shade}{round(100*(#1 - \minvalWeatherMAE)/max(\maxvalWeatherMAE - \minvalWeatherMAE, 0.0001))}%
  \xdef\colorstring{lightblue1!\shade!lightblue2}%
  \cellcolor{\colorstring}{#1}%
}

\newcommand{\heatcellECLMSE}[1]{%
  \pgfmathtruncatemacro{\shade}{round(100*(#1 - \minvalECLMSE)/max(\maxvalECLMSE - \minvalECLMSE, 0.0001))}%
  \xdef\colorstring{lightblue1!\shade!lightblue2}%
  \cellcolor{\colorstring}{#1}%
}
\newcommand{\heatcellECLMAE}[1]{%
  \pgfmathtruncatemacro{\shade}{round(100*(#1 - \minvalECLMAE)/max(\maxvalECLMAE - \minvalECLMAE, 0.0001))}%
  \xdef\colorstring{lightblue1!\shade!lightblue2}%
  \cellcolor{\colorstring}{#1}%
}

\begin{table*}[hbtp!]
\centering
\footnotesize
\renewcommand{\arraystretch}{1.2}
\caption{Long-term forecasting results. Each cell reports (MSE, MAE) averaged over forecasting horizons $H \in \{96, 192, 336, 720\}$. Lower values (indicated by darker shading) are better. While our method is primarily designed to improve ultra-long-term forecasting and performance under data scarcity relative to LLM-only baselines, it is competitive with general long forecasting methods.}
\setlength{\tabcolsep}{3pt}
\resizebox{\textwidth}{!}{%
\begin{tabular}{l p{3pt} cc p{3pt} cc p{3pt} cc p{3pt} cc p{3pt} cc p{3pt} cc p{3pt} cc}
\toprule
\textbf{Method} & & \multicolumn{2}{c}{\textbf{ETTh1}} & & \multicolumn{2}{c}{\textbf{ETTh2}} & & \multicolumn{2}{c}{\textbf{ETTm1}} & & \multicolumn{2}{c}{\textbf{ETTm2}} & & \multicolumn{2}{c}{\textbf{Weather}} & & \multicolumn{2}{c}{\textbf{ECL}} \\
 & & MSE & MAE & & MSE & MAE & & MSE & MAE & & MSE & MAE & & MSE & MAE & & MSE & MAE \\
\midrule
Diffusion-LLM (Ours) &  & \heatcellETThOneMSE{0.427} & \heatcellETThOneMAE{0.446} &  & \heatcellETThTwoMSE{0.387} & \heatcellETThTwoMAE{0.425} &  & \heatcellETTmOneMSE{0.376} & \heatcellETTmOneMAE{0.399} &  & \heatcellETTmTwoMSE{0.334} & \heatcellETTmTwoMAE{0.369} &  & \heatcellWeatherMSE{0.304} & \heatcellWeatherMAE{0.329} &  & \heatcellECLMSE{0.200} & \heatcellECLMAE{0.303} \\
LDM4TS~\cite{ruan2025visionenhancedtimeseriesforecasting} &  &
\heatcellETThOneMSE{0.443} & \heatcellETThOneMAE{0.454} &  &
\heatcellETThTwoMSE{0.387} & \heatcellETThTwoMAE{0.427} &  &
\heatcellETTmOneMSE{0.352} & \heatcellETTmOneMAE{0.387} &  &
\heatcellETTmTwoMSE{0.333} & \heatcellETTmTwoMAE{0.380} &  &
\heatcellWeatherMSE{0.245} & \heatcellWeatherMAE{0.283} &  &
\heatcellECLMSE{0.199} & \heatcellECLMAE{0.299} \\
GPT4TS~\cite{zhou2023one} &  & \heatcellETThOneMSE{0.465} & \heatcellETThOneMAE{0.455} &  & \heatcellETThTwoMSE{0.381} & \heatcellETThTwoMAE{0.412} &  & \heatcellETTmOneMSE{0.388} & \heatcellETTmOneMAE{0.403} &  & \heatcellETTmTwoMSE{0.284} & \heatcellETTmTwoMAE{0.339} &  & \heatcellWeatherMSE{0.237} & \heatcellWeatherMAE{0.270} &  & \heatcellECLMSE{0.167} & \heatcellECLMAE{0.263} \\
DLinear~\cite{zeng_are_2023} &  & \heatcellETThOneMSE{0.422} & \heatcellETThOneMAE{0.437} &  & \heatcellETThTwoMSE{0.431} & \heatcellETThTwoMAE{0.446} &  & \heatcellETTmOneMSE{0.357} & \heatcellETTmOneMAE{0.378} &  & \heatcellETTmTwoMSE{0.267} & \heatcellETTmTwoMAE{0.333} &  & \heatcellWeatherMSE{0.248} & \heatcellWeatherMAE{0.300} &  & \heatcellECLMSE{0.166} & \heatcellECLMAE{0.263} \\
PatchTST~\cite{nie2023a} &  & \heatcellETThOneMSE{0.413} & \heatcellETThOneMAE{0.430} &  & \heatcellETThTwoMSE{0.330} & \heatcellETThTwoMAE{0.379} &  & \heatcellETTmOneMSE{0.351} & \heatcellETTmOneMAE{0.380} &  & \heatcellETTmTwoMSE{0.255} & \heatcellETTmTwoMAE{0.315} &  & \heatcellWeatherMSE{0.225} & \heatcellWeatherMAE{0.264} &  & \heatcellECLMSE{0.161} & \heatcellECLMAE{0.252} \\
TimesNet~\cite{wu2023timesnet} &  & \heatcellETThOneMSE{0.458} & \heatcellETThOneMAE{0.450} &  & \heatcellETThTwoMSE{0.414} & \heatcellETThTwoMAE{0.427} &  & \heatcellETTmOneMSE{0.400} & \heatcellETTmOneMAE{0.406} &  & \heatcellETTmTwoMSE{0.291} & \heatcellETTmTwoMAE{0.333} &  & \heatcellWeatherMSE{0.259} & \heatcellWeatherMAE{0.287} &  & \heatcellECLMSE{0.192} & \heatcellECLMAE{0.295} \\
FEDformer~\cite{pmlr-v162-zhou22g} &  & \heatcellETThOneMSE{0.440} & \heatcellETThOneMAE{0.460} &  & \heatcellETThTwoMSE{0.437} & \heatcellETThTwoMAE{0.449} &  & \heatcellETTmOneMSE{0.448} & \heatcellETTmOneMAE{0.452} &  & \heatcellETTmTwoMSE{0.305} & \heatcellETTmTwoMAE{0.349} &  & \heatcellWeatherMSE{0.309} & \heatcellWeatherMAE{0.360} &  & \heatcellECLMSE{0.214} & \heatcellECLMAE{0.327} \\
Autoformer~\cite{NEURIPS2021_bcc0d400} &  & \heatcellETThOneMSE{0.496} & \heatcellETThOneMAE{0.487} &  & \heatcellETThTwoMSE{0.450} & \heatcellETThTwoMAE{0.459} &  & \heatcellETTmOneMSE{0.588} & \heatcellETTmOneMAE{0.517} &  & \heatcellETTmTwoMSE{0.327} & \heatcellETTmTwoMAE{0.371} &  & \heatcellWeatherMSE{0.338} & \heatcellWeatherMAE{0.382} &  & \heatcellECLMSE{0.227} & \heatcellECLMAE{0.338} \\
Stationary~\cite{liu2022nonstationary} &  & \heatcellETThOneMSE{0.570} & \heatcellETThOneMAE{0.537} &  & \heatcellETThTwoMSE{0.526} & \heatcellETThTwoMAE{0.516} &  & \heatcellETTmOneMSE{0.481} & \heatcellETTmOneMAE{0.456} &  & \heatcellETTmTwoMSE{0.306} & \heatcellETTmTwoMAE{0.347} &  & \heatcellWeatherMSE{0.288} & \heatcellWeatherMAE{0.314} &  & \heatcellECLMSE{0.193} & \heatcellECLMAE{0.296} \\
ETSformer~\cite{woo2023etsformer} &  & \heatcellETThOneMSE{0.542} & \heatcellETThOneMAE{0.510} &  & \heatcellETThTwoMSE{0.439} & \heatcellETThTwoMAE{0.452} &  & \heatcellETTmOneMSE{0.429} & \heatcellETTmOneMAE{0.425} &  & \heatcellETTmTwoMSE{0.293} & \heatcellETTmTwoMAE{0.342} &  & \heatcellWeatherMSE{0.271} & \heatcellWeatherMAE{0.334} &  & \heatcellECLMSE{0.208} & \heatcellECLMAE{0.323} \\
LightTS~\cite{zhang2022morefastmultivariatetime} &  & \heatcellETThOneMSE{0.491} & \heatcellETThOneMAE{0.479} &  & \heatcellETThTwoMSE{0.602} & \heatcellETThTwoMAE{0.543} &  & \heatcellETTmOneMSE{0.435} & \heatcellETTmOneMAE{0.437} &  & \heatcellETTmTwoMSE{0.409} & \heatcellETTmTwoMAE{0.436} &  & \heatcellWeatherMSE{0.261} & \heatcellWeatherMAE{0.312} &  & \heatcellECLMSE{0.229} & \heatcellECLMAE{0.329} \\
Informer~\cite{haoyietal-informer-2021} &  & \heatcellETThOneMSE{1.040} & \heatcellETThOneMAE{0.795} &  & \heatcellETThTwoMSE{4.431} & \heatcellETThTwoMAE{1.729} &  & \heatcellETTmOneMSE{0.961} & \heatcellETTmOneMAE{0.734} &  & \heatcellETTmTwoMSE{1.410} & \heatcellETTmTwoMAE{0.810} &  & \heatcellWeatherMSE{0.634} & \heatcellWeatherMAE{0.548} &  & \heatcellECLMSE{0.311} & \heatcellECLMAE{0.397} \\
Reformer~\cite{kitaev2020reformerefficienttransformer} &  & \heatcellETThOneMSE{1.029} & \heatcellETThOneMAE{0.805} &  & \heatcellETThTwoMSE{6.736} & \heatcellETThTwoMAE{2.191} &  & \heatcellETTmOneMSE{0.799} & \heatcellETTmOneMAE{0.671} &  & \heatcellETTmTwoMSE{1.479} & \heatcellETTmTwoMAE{0.915} &  & \heatcellWeatherMSE{0.803} & \heatcellWeatherMAE{0.656} &  & \heatcellECLMSE{0.338} & \heatcellECLMAE{0.422} \\
\bottomrule
\end{tabular}
}
\label{tab:forecasting_results_reordered}
\end{table*}

\section{Model Analysis}

We present ablation studies highlighting design choices, with empirical results in Table~\ref{tab:etth1_ablation_ultra_long} (Appendix~\ref{subsec:expt_details}).

\noindent\textbf{Architectural Variants and Conditioning Strategies:}  
We compared a 1D U-Net~\cite{10.1007/978-3-319-24574-4_28} with a fully connected DDPM. Despite U-Net’s capacity, the simpler architecture performed similarly or better, indicating overparameterization is unnecessary.

DDPM conditions on concatenated prompt and time-series embeddings. We also tested concatenation versus attention-based conditioning and found simple concatenation most robust (A.1., A.3. in Table~\ref{tab:etth1_ablation_ultra_long}).

\noindent\textbf{Channel Independence:}  
Adding feature-ID conditioning slightly degraded performance (A.1., A.2. in Table~\ref{tab:etth1_ablation_ultra_long}), suggesting DDPM benefits from shared latent representations rather than explicit channel separation.

\noindent\textbf{Encoder Sharing and DDPM Contribution:}  
DDPM with separate encoders for lookback and forecast improved ultra-long-term ETTh1 forecasting by \(10.81\%\); shared encoder added another \(12.48\%\) gain (A.1., B.1., B.2. in Table~\ref{tab:etth1_ablation_ultra_long}). Figure~\ref{fig_lambda_mse} shows the effect of $\lambda$ with best performance at $\lambda=1$, balancing LLM and DDPM contributions.

\begin{figure}[tb]
\centering
\resizebox{0.85\linewidth}{!}{
\begin{subfigure}[t]{0.45\textwidth}
    \centering
    \includegraphics[width=\linewidth]{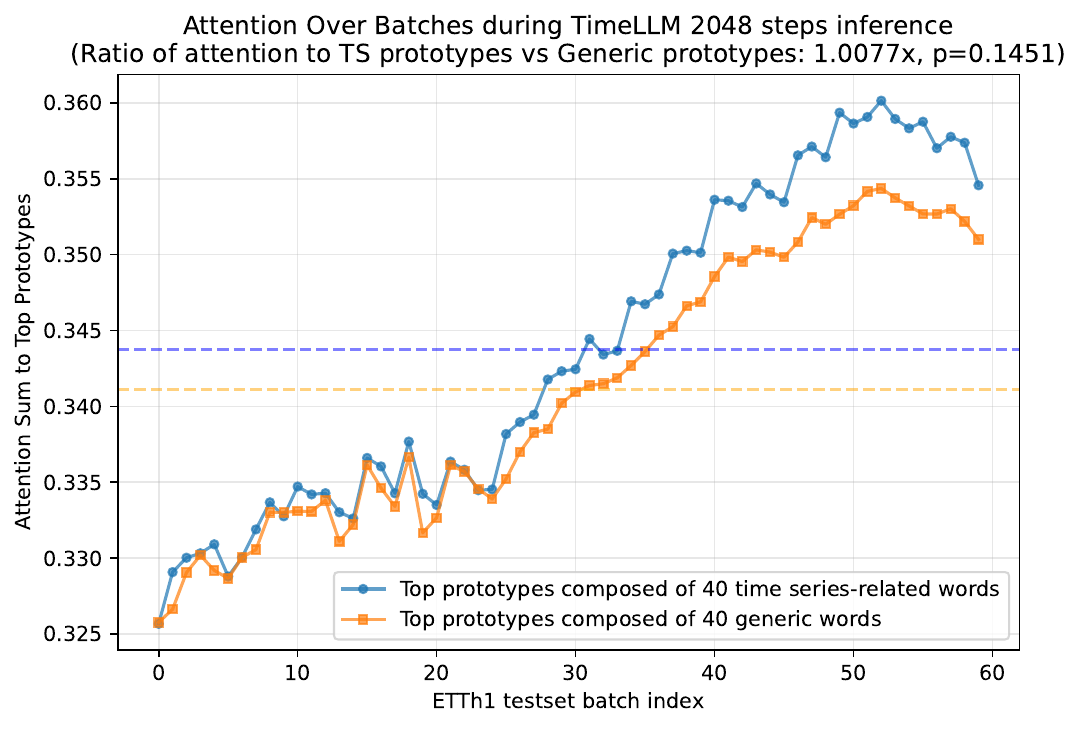}
    \caption{TimeLLM}
    \label{fig:proto_att_timellm}
\end{subfigure}
\hfill
\begin{subfigure}[t]{0.45\textwidth}
    \centering
    \includegraphics[width=\linewidth]{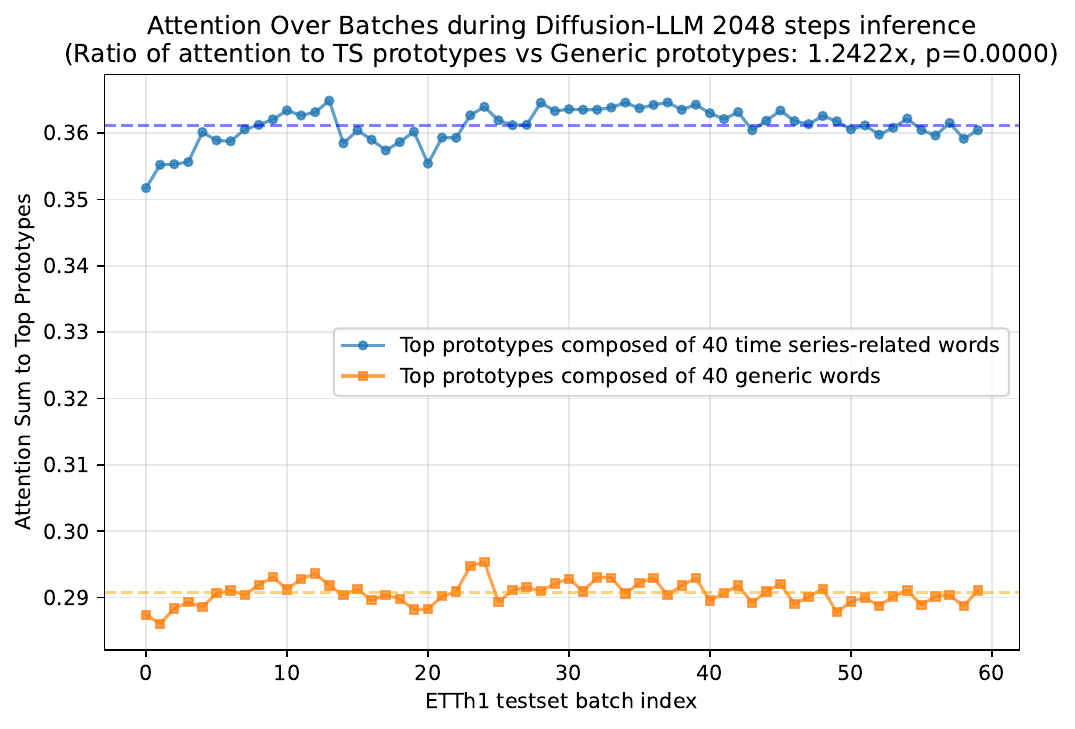}
    \caption{\texttt{Diffusion-LLM}}
    \label{fig:proto_att_diffllm}
\end{subfigure}
}
\caption{We derive the top 10 prototypes for time-series-related words and for generic words (like figure~\ref{fig_proto_compose}) for 2048 forecast inference on ETTh1. Then we plot the attention allocated over these two types of prototypes. \texttt{Diffusion-LLM} shows significantly stronger attention toward time-series prototypes, indicating improved alignment between time series patches and semantic prototypes.}
\label{fig:proto_attention}
\end{figure}

\noindent\textbf{Multimodal Alignment Analysis:}
Figure~\ref{fig:proto_attention} shows \texttt{Diffusion-LLM} places noticeably stronger attention on time-series-related prototypes than TimeLLM, indicating more stable temporal–semantic alignment under extreme horizons.

\noindent\textbf{Efficiency Analysis:}  
Compared to TimeLLM, Diffusion-LLM adds minimal overhead in the largest setting (2048 steps): +1.82\% GPU memory, +11.54\% parameters, and only 0.39\% slower training (Table~\ref{tab:efficiency_analysis}). Inference uses only LLM (Figure~\ref{fig_infer}), so speed remains unchanged.

\section{Conclusion}

In this work, we introduced \texttt{Diffusion-LLM}, a low-overhead but powerful extension to LLM-based time series forecasting frameworks that integrates a conditional diffusion model for distributional regularization. Our method improves performance in ultra-long-term forecasting and few-shot learning scenarios, where uncertainty and data scarcity pose major challenges. By modeling the conditional distribution of future representations in the shared embedding space, \texttt{Diffusion-LLM} enhances the LLM's ability to reason over long horizons and generalize from limited data. 
Promising future directions include more adaptive reprogramming strategies, applying diffusion-based regularization to other embedding spaces, and exploring diffusion for direct generative forecasting to enable uncertainty-aware multi-predictions. Extending the framework for additional modalities or for LLM uncertainty estimation  also remains an exciting direction. 
\texttt{Diffusion-LLM} offers a principled and effective enhancement to time series LLMs, combining the strengths of probabilistic modeling and pretrained language models in a unified framework without loss of existing efficiency.

\noindent\textbf{Acknowledgements. }This research was made possible through an industry collaboration with the Audi PhD Program. We also acknowledge HPC resources from NHR@FAU (projects b143dc, b180dc), funded by federal and Bavarian state authorities and Gerhard Wellein's and his team's HPC approach. NHR@FAU hardware is partially funded by DFG 440719683. Additional support was received from ERC projects MIA-NORMAL 101083647, DFG 513220538 and 512819079, and the state of Bavaria (HTA). We used coding agents and LLMs from Anthropic, OpenAI, Google, and Mistral AI, for text polishing, coding, experiment orchestration, and cluster monitoring.


%
%
%
\bibliographystyle{splncs04}
\bibliography{icann2026_conference}

\clearpage
\appendix
\pagenumbering{roman}

\appendix

\section{Supplementary Material}

\subsection{Dataset Details}
\label{subsec:dataset_details}

We evaluate \textbf{Diffusion-LLM} on six widely-used benchmark datasets for long-term time series forecasting. These datasets span multiple domains, including energy, weather, and offer a diverse testbed for assessing the performance and generalization of our method. The ILI dataset~\cite{wu2023timesnet} was considered but its shorter standard forecast window of $H \in \{24, 36, 48, 60\}$ and the unavailability of enough data for ultra-long forecasting make it unsuitable for our evaluation.

\begin{itemize}
    \item \textbf{ETTm1 and ETTm2}: These datasets are derived from the Electricity Transformer Temperature (ETT) dataset. ETTm1 and ETTm2 contain measurements sampled every 15 minutes, with seven features including oil temperature and load.

    \item \textbf{ETTh1 and ETTh2}: These datasets also come from the ETT collection but are sampled at an hourly resolution. Like ETTm1 and ETTm2, it includes seven variables, capturing environmental and operational characteristics of electric transformers.

    \item \textbf{Weather}: The Weather dataset is sourced from the UCI Machine Learning Repository and contains meteorological data collected from a local weather station. It includes 21 continuous variables (e.g., temperature, humidity, pressure) recorded every 10 minutes.

    \item \textbf{ECL (Electricity Consumption Load)}: This dataset consists of hourly electricity consumption data from 321 clients in Europe.
\end{itemize}

\begin{table*}[h]
\centering
\footnotesize
\renewcommand{\arraystretch}{1.2}
\setlength{\tabcolsep}{6pt}
\resizebox{\textwidth}{!}{%
\begin{tabular}{lcccccc}
\toprule
\textbf{Dataset} & \textbf{Dim.} & \textbf{Dataset Size (Train, Val, Test)} & \textbf{Frequency} & \textbf{Domain} & \textbf{Task} \\
\midrule
ETTm1       & 7   & (34465, 11521, 11521) & 15 min  & Temperature   & Long-term Forecasting \\
ETTm2       & 7   & (34465, 11521, 11521) & 15 min  & Temperature   & Long-term Forecasting \\
ETTh1       & 7   & (8545, 2881, 2881)    & 1 hour  & Temperature   & Long-term Forecasting \\
ETTh2       & 7   & (8545, 2881, 2881)    & 1 hour  & Temperature   & Long-term Forecasting \\
Weather     & 21  & (36792, 5271, 10540)  & 10 min  & Weather       & Long-term Forecasting \\
Electricity & 321 & (18317, 2633, 5261)   & 1 hour  & Electricity   & Long-term Forecasting \\
\bottomrule
\end{tabular}%
}
\caption{Overview of datasets used in Diffusion-LLM. Each dataset varies in dimensionality, sampling frequency, and domain. Forecasting horizons are standardized across all datasets.}
\label{tab:dataset_overview}
\end{table*}

For all datasets, we follow the standard data preprocessing and splitting protocols used in prior work such as PatchTST and Time-LLM (Available from the library in \url{https://github.com/thuml/Time-Series-Library/tree/main}). Specifics of the dataset are added in table \ref{tab:dataset_overview}.

\subsection{Evaluation Metrics}

To evaluate model performance on time series forecasting, we adopt two standard regression metrics:

\begin{itemize}
    \item \textbf{Mean Squared Error (MSE)}: This metric computes the average of the squared differences between the predicted values and the ground truth:
    \[
    \text{MSE} = \frac{1}{N} \sum_{i=1}^{N} (y_i - \hat{y}_i)^2
    \]
    A lower MSE indicates better performance and penalizes larger errors more heavily due to the squared term.

    \item \textbf{Mean Absolute Error (MAE)}: MAE measures the average absolute difference between predictions and actual values:
    \[
    \text{MAE} = \frac{1}{N} \sum_{i=1}^{N} |y_i - \hat{y}_i|
    \]
    MAE is more robust to outliers compared to MSE and provides an intuitive measure of forecast accuracy.
\end{itemize}

\subsection{Experiment Details}
\label{subsec:expt_details}

\noindent\textbf{Model Architecture:}

Our model adopts a denoising diffusion probabilistic modeling (DDPM) framework for time series forecasting. The underlying structure is a lightweight residual multilayer perceptron (MLP). The model consists entirely of fully connected layers and skip connections.

Let \( x \in \mathbb{R}^{B \times L \times D} \) denote a batch of input time series, where \( B \) is the batch size, \( L \) is the sequence length, and \( D \) is the input dimensionality. The model maps a noisy input \( x_t \) to a denoised prediction \( \hat{x}_0 \) through the following components:

\noindent\textbf{Input and Context Projection:} 

The input sequence is projected from \( D \) to a hidden dimension \( H \) via a linear layer. A conditioning signal (e.g., a context window or past data), also of dimension \( D \), is mean-pooled over the temporal axis, broadcast to match the sequence length, and projected into the same hidden space. The two are summed along with a time embedding to produce the initial hidden state:
\[
h = \text{Linear}_\text{in}(x) + \text{Linear}_\text{cond}(\text{repeat}(\text{mean}(c))) + \text{TimeEmbedding}(t)
\]

\noindent\textbf{Time Embedding:} 

To encode the diffusion timestep \( t \), we use a sinusoidal embedding of dimension \( H \), similar to positional embeddings in transformers. This embedding is passed through a linear layer and ReLU activation:
\[
t_{\text{emb}} = \text{ReLU}(\text{Linear}_\text{time}(\text{Sinusoidal}(t)))
\]
This time embedding is broadcast across the temporal dimension and added to the hidden state.

\noindent\textbf{Class Conditioning (Optional):} 

The different features in the dataset are used as different classes for the conditional DDPM. Each class is added to the hidden representation at every timestep.

\noindent\textbf{Residual Blocks:} 
The hidden representation is passed through two residual blocks, each consisting of a linear layer followed by a GELU activation and residual skip connection:
\[
h \leftarrow h + \text{GELU}(\text{Linear}(h))
\]

\noindent\textbf{Output Projection:} 

Finally, a linear output layer maps the hidden representation back to the original input dimension:
\[
\hat{x}_0 = \text{Linear}_\text{out}(h)
\]

\noindent\textbf{Noise Schedule:}

We experiment with two types of noise schedules for the diffusion process:

\begin{itemize}
    \item \textbf{Linear Schedule.} A simple linear beta schedule is defined as:
    \[
    \beta_t = \text{linspace}\left( \frac{1000}{T} \cdot 10^{-4}, \frac{1000}{T} \cdot 0.02, T \right)
    \]
    where \( T \) is the total number of diffusion steps.

    \item \textbf{Cosine Schedule.} We define the cosine schedule over \( T \) steps as:
    \[
    \bar{\alpha}_t = \cos^2\left( \frac{(t/T + s)}{1 + s} \cdot \frac{\pi}{2} \right), \quad \beta_t = 1 - \frac{\bar{\alpha}_{t+1}}{\bar{\alpha}_t}
    \]
    where \( s \) is a small constant (e.g., \( 0.008 \)), and \( \beta_t \) is clipped to the range \([0, 0.999]\) for numerical stability.
\end{itemize}

\begin{table}[t]
\centering
\scriptsize
\renewcommand{\arraystretch}{1.05}
\setlength{\tabcolsep}{3pt}
\begin{tabular}{lc}
\toprule
\multicolumn{1}{c}{\textbf{Variant}} & \textbf{ETTh1-2048 MSE} \\ 
\midrule
\textbf{A.1.} DiffusionLLM & \textbf{0.729} \\
\textbf{A.2.} DiffusionLLM with Class Conditioning (A2) & 0.746 \\
\textbf{A.3.} DiffusionLLM with Complex U-Net \& Attention Conditioning & 0.732 \\
\textbf{B.1.} DDPM with Separate Lookback and Forecast Encoders & 0.833 \\
\textbf{B.2.} Without DDPM (TimeLLM-style baseline) & 0.934 \\
\bottomrule
\end{tabular}
\caption{\added[id=anon]{Ablations on ETTh1 in predicting 2048 steps ahead (MSE reported). Best result highlighted in bold.}}
\label{tab:etth1_ablation_ultra_long}
\end{table}

\begin{table*}[t]
\centering
\footnotesize
\renewcommand{\arraystretch}{1.2}
\setlength{\tabcolsep}{4pt}
\resizebox{\textwidth}{!}{%
\begin{tabular}{lcccccccccc}
\toprule
\textbf{Task-Dataset} & \textbf{Text Prototype} & \textbf{Backbone Layers} & \textbf{Input Length $T$} & \textbf{Patch Dim. $d_m$} & \textbf{Heads $K$} & \textbf{FF Dim. $d_{ff}$} & \textbf{LR$^*$} & \textbf{Loss} & \textbf{Batch Size} & \textbf{Epochs} \\
\midrule
LTF - ETTh1     & 1000 & 32 & 512 & 16 & 8 & 128 & $10^{-3}$ & MSE & 16  & 50  \\
LTF - ETTh2     & 1000 & 32 & 512 & 16 & 8 & 128 & $10^{-3}$ & MSE & 16  & 50  \\
LTF - ETTm1     & 1000 & 32 & 512 & 16 & 8 & 128 & $10^{-3}$ & MSE & 16 & 100 \\
LTF - ETTm2     & 1000 & 32 & 512 & 16 & 8 & 128 & $10^{-3}$ & MSE & 16 & 100 \\
LTF - Weather   & 1000 & 32 & 512 & 16 & 8 & 128 & $10^{-2}$ & MSE & 64  & 100 \\
LTF - ECL       & 1000 & 32 & 512 & 16 & 8 & 32  & $10^{-2}$ & MSE & 128 & 100 \\
\bottomrule
\end{tabular}%
}
\caption{LLM hyperparameters used for each dataset in Diffusion-LLM. All models use the same LLaMA-7B backbone with frozen weights.}
\label{tab:llm_hyperparams}
\end{table*}

\begin{table*}[!htbp]
\centering
\footnotesize
\renewcommand{\arraystretch}{1.0}
\setlength{\tabcolsep}{5pt}
\resizebox{\textwidth}{!}{%
\begin{tabular}{l l}
\toprule
\textbf{Hyperparameter} & \textbf{Value / Description} \\
\midrule
\texttt{input\_dim} & 4096 (Dimensionality of input time series patches) \\
\texttt{hidden\_dim} & 512 (Hidden layer size used throughout the DDPM model) \\
\texttt{time\_emb\_dim} & 512 (Dimensionality of sinusoidal time embedding) \\
\texttt{num\_classes} & 0 (No class conditioning used in final version) \\
\texttt{residual\_blocks} & 2 (Number of residual blocks in the DDPM architecture) \\
\texttt{activation} & GELU (Activation function used in residual blocks) \\
\texttt{output\_proj} & Linear (Final layer to project hidden state back to input dimension) \\
\texttt{timesteps} & 1000 (Total number of diffusion steps) \\
\texttt{beta\_schedule} & cosine (Noise schedule used for diffusion process) \\
\texttt{sampling\_timesteps} & 1000 (Number of steps used during sampling) \\
\texttt{objective} & pred\_noise (Training objective: predict added noise) \\
\texttt{loss\_function} & MSE (Loss computed between predicted and target noise) \\
\texttt{self\_conditioning} & False (Optional technique to improve sample quality; not used) \\
\texttt{parameter\_count} & $\sim$7M (Approximate number of parameters added by DDPM) \\
\bottomrule
\end{tabular}%
}
\caption{DDPM hyperparameters used in Diffusion-LLM. These settings are shared across all datasets.}
\label{tab:ddpm_hyperparams}
\end{table*}

To avoid underestimating our baseline, for the LLM part, we use the same hyperparameters as \cite{jin2024timellm} apart from Weather and Electricity dataset where we use larger batch size of 64 and 128 to accommodate computing time. The hyperparameters are listed in the table \ref{tab:llm_hyperparams}.

For our DDPM architecture, we use same hyperparameters for all datasets. It is a residual MLP-based backbone with a hidden dimension of 512 throughout. The input and conditioning sequences, each with dimensionality 4096, are projected to the hidden space using separate linear layers. The model includes two residual blocks, each with a single linear layer followed by GELU activation and skip connection. A sinusoidal time embedding of size 512 is used, followed by a linear projection to match the hidden dimension. The output is projected back to the original input dimension via a final linear layer. Overall, the model contains six main linear layers, all operating at the hidden size of 512. The DDPM model adds only approximately 7 M parameters. Further, adding the condition into the DDPM model in different scenarios for different datasets always yielded similar results with 1-2 percent deviation only in either direction, hence in the final version, we have not used the class conditioning. The DDPM hyperparameters are listed in the table \ref{tab:ddpm_hyperparams}.

\begin{figure*}[t]
\centering
\includegraphics[width=0.9\textwidth]{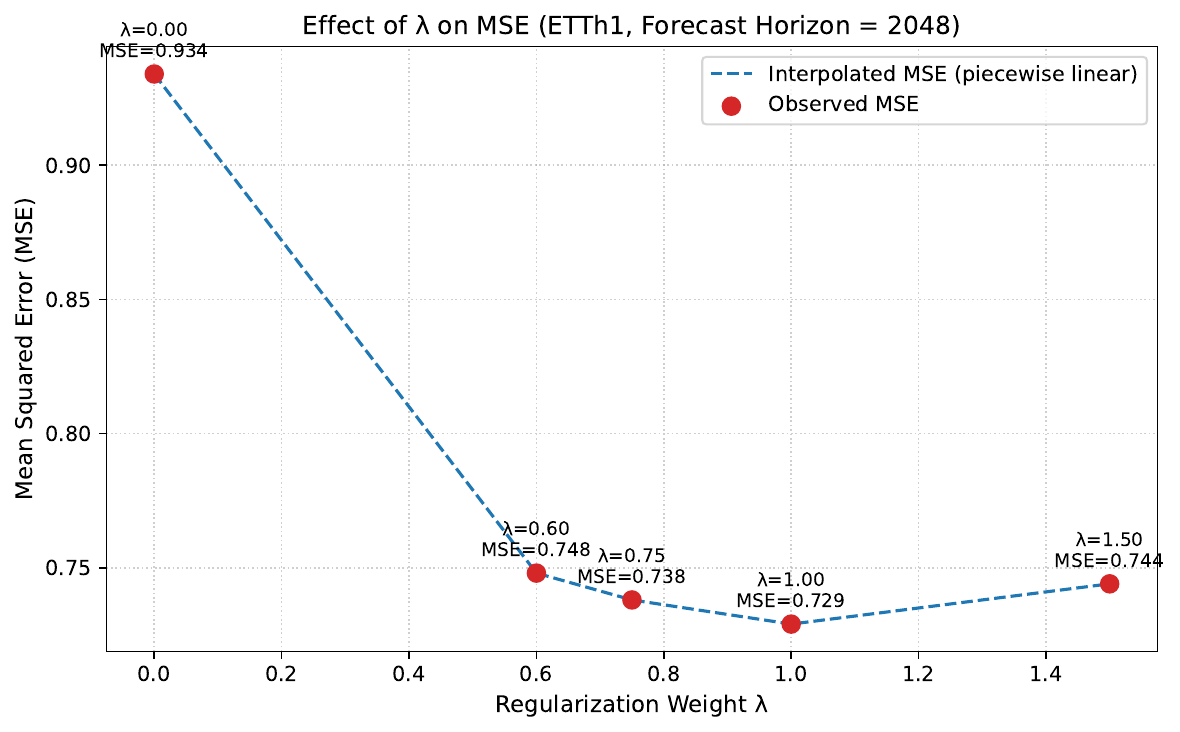} 
\caption{\added[id=anon]{Impact of regularization weight ($\lambda$) on forecasting performance (MSE) for ETTh1 dataset with a 2048-step horizon. The plot shows that $\lambda$ = 1 achieves the best performance (MSE = 0.729), indicating that an equal contribution from the forecasting loss and the diffusion-based regularization provides optimal balance. Smaller $\lambda$ values (e.g., 0 for TImeLLM or 0.6) under-regularize the embedding space, limiting the benefit of distribution-aware alignment, while larger $\lambda$ values (e.g., 1.5) overemphasize the diffusion objective, causing over-regularization and slight performance degradation. This demonstrates the importance of tuning $\lambda$ to balance deterministic forecasting and probabilistic embedding refinement.}}
\label{fig_lambda_mse}
\end{figure*}

\begin{table}[t]
\centering
\renewcommand{\arraystretch}{1.15}  
\setlength{\tabcolsep}{6pt}         
\resizebox{\linewidth}{!}{%
\begin{tabular}{lcccc}
\toprule
\textbf{Model} & \textbf{Training Time (GPU-h)} & \textbf{Max GPU Mem Usage (MiB)} & \textbf{Trainable Params (M)} & \textbf{Speed (s/iter)} \\
\midrule
Diffusion-LLM & 6.437 & 33188 & 6.461 & 0.397 \\
TimeLLM       & 6.461 & 32592 & 6.437 & 0.395 \\
\bottomrule
\end{tabular}
}
\caption{\added[id=anon]{Efficiency analysis for ETTh1 forecasting 2048 steps ahead. Training time and resource usage are reported for Diffusion-LLM and TimeLLM.}}
\label{tab:efficiency_analysis}
\end{table}


\end{document}